%% file: nips_2017.tex
\documentclass{article}

\usepackage[final]{nips_2017}

\usepackage[utf8]{inputenc} %
\usepackage[T1]{fontenc}    %
\usepackage{hyperref}       %
\usepackage{url}            %
\usepackage{booktabs}       %
\usepackage{amsmath}        
\usepackage{amsfonts}       %
\usepackage{nicefrac}       %
\usepackage{microtype}      %
\usepackage{graphicx}
\usepackage{xcolor}
\usepackage{setspace}

\include{newcom}

\include{macros}

\usepackage{natbib}
\title{Grounding Visual Explanations (Extended Abstract)}

\author{
  Lisa Anne Hendricks$^1$,  Ronghang Hu$^1$, Trevor Darrell$^1$ and Zeynep Akata$^2$\\
  $^1$ EECS, University of California Berkeley \\
  $^2$ Amsterdam Machine Learning Lab, University of Amsterdam
}

\begin{document}

\maketitle

\begin{abstract} 
Existing models~\cite{hendricks16eccv}  which generate textual explanations enforce task relevance through a discriminative term loss function, but such mechanisms only weakly constrain mentioned object parts to actually be present in the image.
In this paper, a new model is proposed for generating explanations by utilizing localized grounding of constituent phrases in generated explanations to ensure image relevance.  
Specifically, we introduce a phrase-critic model to refine (re-score/re-rank) generated candidate explanations and employ a relative-attribute inspired ranking loss using `flipped' phrases as negative examples for training. 
At test time, our phrase-critic model takes an image and a candidate explanation as input and outputs a score indicating how well the candidate explanation is grounded in the image.  
\end{abstract}

\section{Introduction}

Modern neural networks are good at localizing objects, predicting object categories and describing scenes with natural language. 
However, the decision processes of neural networks are often opaque.  
Therefore, in order to interpret and monitor neural networks, providing explanations of network decisions has gained interest.
Here, we focus on providing such explanations via natural language, i.e. textual explanations.
A textual explanation system for a classification network ideally both discusses discriminative features of its predicted classes and names image-relevant attributes.
However, sometimes these two goals are in opposition.
For example, if one discriminative attribute frequently occurs within a class, a model may learn to justify its prediction by mentioning this attribute as a discriminative feature even if the input does not contain the attribute.
In this work, we aim at resolving such conflicts and design a framework that automatically generates textual explanations that justify a classification decision and simultaneously ground discriminative object properties both in the explanation and in the image via a novel a phrase-critic model. 
Our model significantly improves the image-relevance of explanations in comparison to prior works.

In our framework, a phrase-critic model is first trained specifically to ground phrases, irrespective of linguistic fluency. 
Fluency is ensured by training an LSTM-based explanation model to generate candidate sentences which discuss class discriminative features.
In other words, since our phrase-critic model does not focus on fluency, it should be more reliable when understanding sentence correctness; meanwhile, our explanation generation mechanism ensures the appearance of class discriminative information as well as fluency.
As a result, we obtain more accurate and linguistically satisfying explanations than those generated by only enforcing a discriminative training term as done in prior work~\cite{hendricks16eccv}.
An important side effect of our method is visual explanation--a visualization of the grounding of discriminative object parts in the image that are mentioned in the explanation.

\section{Related Work}

In \cite{teach1981analysis}, trust is regarded as a primary reason to explore explainable intelligent systems.
We argue a system which outputs discriminative features of an object class without being image relevant is likely to lose the trust of users.
Consequently, we seek to explicitly enforce image relevancy with our model.

Similar to \cite{biran2014justification}, we aim at providing \textit{justifications} to explain which evidence is important for a decision as opposed to introspective explanations that explains the intermediate activations of neural networks.
Recently, \cite{hendricks16eccv} proposed to generate natural language justifications of a fine-grained object classifier. 
However, it does not ground the relevant object parts in the sentence or the image. 
In \cite{park2016attentive}, similar explanations are generated for activities and VQA pairs. 
Although an attention based explanation system is proposed, there are no constraints to ensure the actual presence of the mentioned attributes or entities in the image. 
Consequentially, these related works~\cite{hendricks16eccv,park2016attentive}, albeit generating convincing textual explanations, do not include a process for networks to correct themselves if their textual explanation is not well-grounded visually.
In contrast, we propose a general process to first check whether explanations are accurately aligned with image input and then improve textually explanations by selecting a better-aligned candidate.

\begin{figure}[t]
\centering
\vspace{-1cm}
\includegraphics[width=0.9\linewidth]{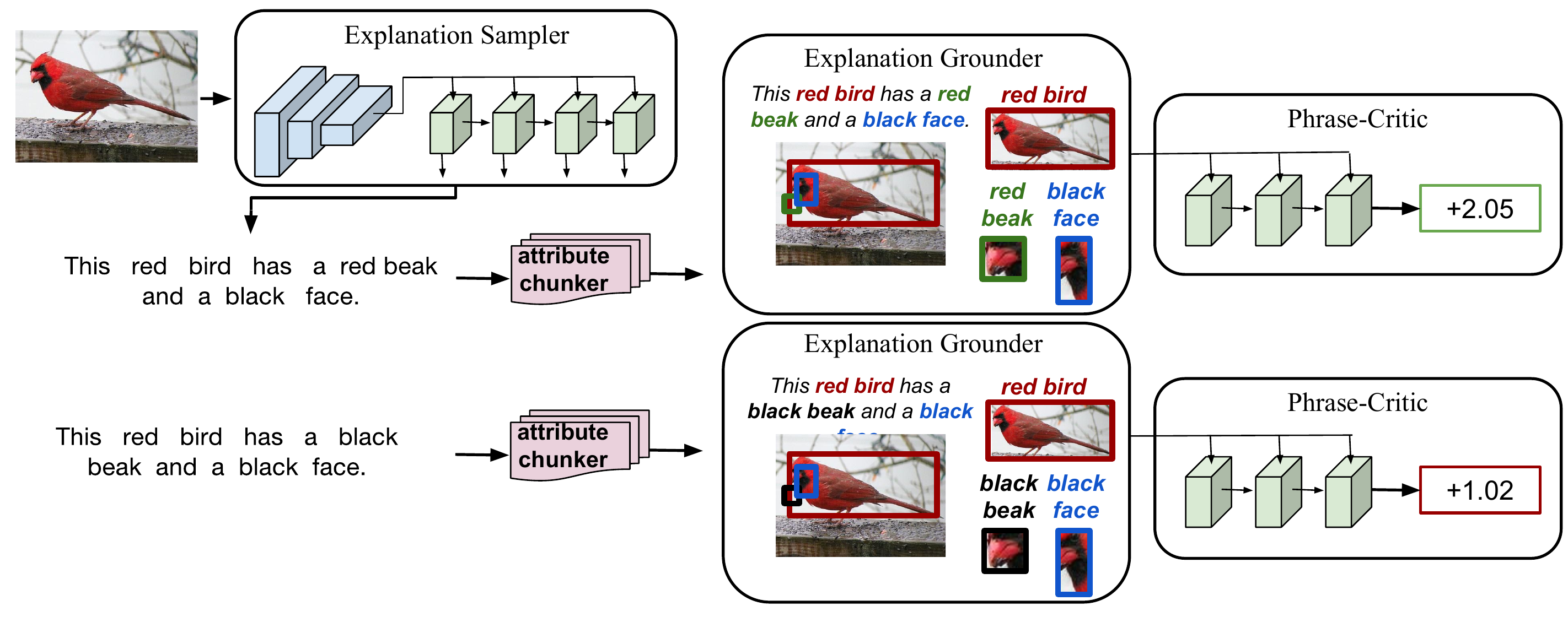}
\caption{The proposed model incorporates a phrase-critic to ensure that generated explanations are not only class discriminative, but also image relevant. We first sample a set of explanations using the generator from~\cite{hendricks16eccv}, then visually ground constituent phrases using a grounder based on~\cite{hu2017modeling}. Then, our phrase-critic model assigns a score to each explanation which reflects whether the explanation is image relevant.
Our model encourages higher ranked sentences to be both class discriminative and image relevant.}
\label{fig:teaser}
\end{figure}

\section{Grounded Visual Explanations}

Our model consists of three main components.
First, a text generation system (based on \cite{hendricks16eccv}) samples many (100 in our experiments) textual explanations.
\cite{hendricks16eccv} is trained with a discriminative loss to encourage sentences to mention class specific attributes.
Next, a phrase grounding model (based on \cite{hu2017modeling}) grounds phrases in the generated textual explanations.
Finally, our proposed phrase-critic model determines which textual explanations are preferred based on how well textual explanations are grounded in the image.
As shown in Figure~\ref{fig:teaser}, our system first generates possible explanations (e.g., ``The red bird has a red beak and a black face''), determines whether constituent phrases (e.g., ``red bird'', ``red beak'',``black face'') are present in the image, and then assigns a score to the explanation.

\textbf{Phrase-Critic.} Our phrase-critic model is the core to our framework. 
It takes a list of  $\{(\mathcal{A}_i, \mathcal{R}_i, s_i)\}$, where $\mathcal{A}_i$ is the attribute phrase, $\mathcal{R}_i$ is the corresponding region (more precisely, visual features extracted from the region), and $s_i$ the region score, and maps them into a single image relevance score $S_r$. For a given attribute phrase $A_i$ such as ``black beak'', we ground (localize) it into a corresponding image region $R_i$ and obtain its localization score $s_i$, using an out-of-shelf localization model from \cite{hu2017modeling} pretrained on VisualGenome.  
It is worth noting that the scores directly produced by the grounding model are not comparable across images and difficult to be directly combined with other metrics, such as sentence fluency, because these scores are difficult to normalize across different images and different visual parts.
For example, a correctly grounded phrase ``yellow belly'' may have a much smaller score than the correctly grounded phrase ``yellow eye'' because a bird belly is less well defined than a bird eye; for another example, an occluded image tend to score lower with all explanations 
Henceforth, our phrase-critic model plays an essential role in producing normalized, utilizable and comparable scores. 
More specifically, given an image $I$, the phrase-critic model processes the list of $\{(\mathcal{A}_i, \mathcal{R}_i, s_i)\}$ by first encoding each $(A_i, \mathcal{R}_i, s_i)$ into a fixed-dimensional vector $x_{enc}$ with an LSTM and then applying a two-layer neural network to regress the final score $S_r$ which reflects the overall image relevancy of an explanation.

\begin{equation}
S_r (\{A_i\}, I; \theta) = f_{critic}(\{(A_i, \mathcal{R}_i, s_i)\}; \theta)
\end{equation}
where $f_{critic}$ consists of an LSTM encoder $f_{lstm}$ and a small two-layer network $f_{nn}$.

We construct a few explanation pairs for each image. Each explanation pair consists of a positive explanation (image-relevant) and a negative explanation (not image-relevant). We then train our explanation critic using the following margin-based ranking loss $\Loss_\mathrm{rank}$ on each pair of positive and negative explanations, to encourage the model to give higher scores to positive explanations than negative explanations:
\begin{align}
\Loss_\mathrm{rank}
& = \max(0, S_r(\{A_{i}^{neg}\}, I; \theta) - S_r(\{A_{i}^{pos}\}, I; \theta) + M)
\end{align}
where $\{A_{i}^{pos}\}$ are matching attribute phrase whereas $\{A_{i}^{neg}\}$ are mismatching attribute phrases respectively, therefore $S_{r}^{pos}$ and $S_{r}^{neg}$ are the scores of the positive and the negative explanations. We use $M=1$ in our implementation.
 
\textbf{Flipped Attribute Training.} The simplest way to sample a negative pair is to consider a mismatching ground truth image and sentence pair.
However, due to the fine-grained nature of our dataset, we empirically found that naively sampling out-of-class negative examples can risk the negative examples being visually too different (such as a Cardinal and an American Crow).
Inspired by a relative attribute paradigm for recognition and retrieval \cite{relativeattributesiccv}, we create negative examples by flipping attributes corresponding to color and size in attribute phrases.  
For example, if a ground truth sentence mentions a ``yellow belly'' and ``red head'' we might change the attribute phrase ``red head'' to ``black head''.  
This means the negative explanation still mentions some attributes present in the image, but is not completely correct.

\textbf{Ranking Explanations.}
After generating a set of candidate explanations and extracting an explanation score using our explanation model, we choose the best explanation based on the score for each explanation.
In practice, we find it is important to rank sentences based on both the relevance score $S_r$ and a fluency score $S_f$ (defined as the $\log P(w_{0:T})$). 
Including $S_f$ is important because the explanation scorer will rank ``This is a bird with a long neck, long neck, and red beak'' high (if a long neck and red beak are present) even though mentioning ``long neck'' twice is clearly ungrammatical.

\textbf{Grounding Visual Features.} The framework for grounding visual features involves three steps: generating visual explanations, factorizing the sentence into smaller chunks, and localizing each chunk with a grounding model. 
Visual explanations are produced using the model of \cite{hendricks16eccv}.
For each explanation we extract a list of $i$ attribute phrases ($\mathcal{A}_i$) using a rule-based attribute phrase chunker.

Once we have extracted attribute phrases $\mathcal{A}_i$, we ground each of them to a visual region $\mathcal{R}_i$ in the original image by using the baseline model proposed in \cite{hu2017modeling} trained on the Visual Genome dataset \cite{krishna2016visual}.
For a given attribute phrase $\mathcal{A}_i$, the grounding model localizes the phrase into an image region, returning a bounding box $\mathcal{R}_i$ and a score $s_i$ of how likely the returned bounding box matches the phrase. 
The attribute phrase, the corresponding region, and the region score form an attribute phrase grounding $(\mathcal{A}_i, \mathcal{R}_i, s_i)$.
This attribute phrase grounding is used as an input to our phrase-critic.

Whereas visual descriptions are encouraged to discuss attributes which are relevant to a specific class, the grounding model is only trained to determine whether a natural language phrase is in an image. Being discriminative rather than generative, the critic model does not have to learn to generate fluent, grammatically correct sentences, and can thus focus on checking whether the mentioned attribute phrases are image-relevant. Consequently, the models are complementary, allowing one model to catch the mistakes of the other.

\section{Experiments}
\begin{figure}[t]
\centering
\vspace{-1cm}
\includegraphics[width=0.9\linewidth]{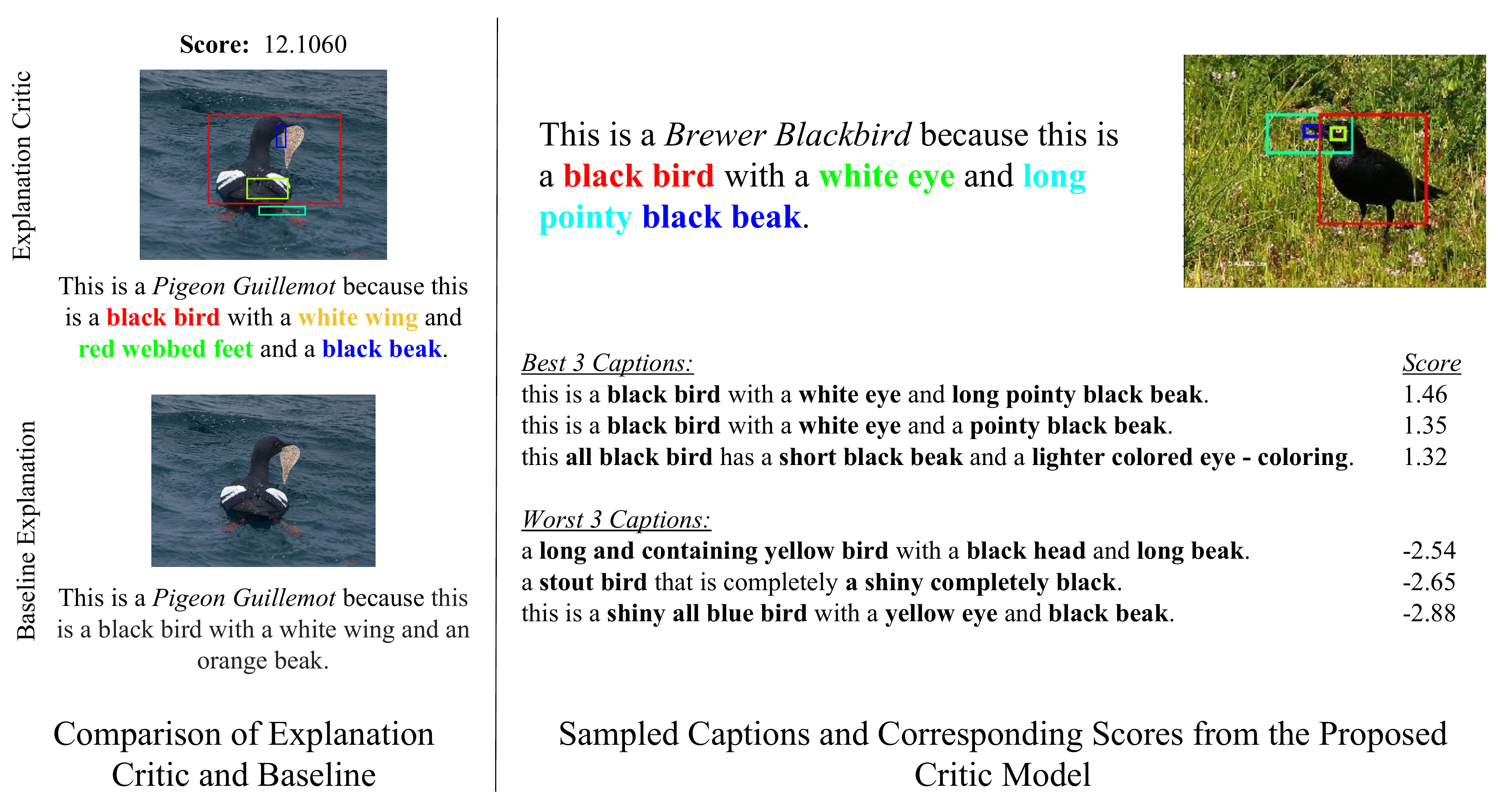}
\caption{(Left) Compare our explanation critic to the baseline model~\cite{hendricks16eccv}. For our model, detected bounding boxes are color coded with their matching attribute phrases in the sentence. The cumulative matching score of the explanation sentence and the visually grounded bounding box are provided above the image.  (Right) We show examples of sampled sentences and corresponding scores from the ranker.  Sentences which are more image relevant are scored higher by our phrase-critic model.}
\label{fig:CriticVBaseline}
\vspace{-2mm}
\end{figure}

For our experiments, we use the Caltech UCSD Birds 200-2011 (CUB) dataset~\cite{welinder10tr} and sentences collected by ~\cite{RALS16}.
We first compare our proposed model with the baseline visual explanation model of~\cite{hendricks16eccv}. 
We present results in Figure~\ref{fig:CriticVBaseline}. 
As a general observation, our critic model (1) grounds attribute phrases both in the image and in the sentence, (2) is in favor of accurate and class-specific attribute phrases and (3) provides the cumulative score of each explanatory sentence.  
To further emphasize the importance of grounding attribute phrases in the image and in the sentence in evaluating the accuracy of the visual explanation model, let us more closely examine Figure~\ref{fig:CriticVBaseline} left.
We note that the baseline model mentions an ``orange beak''.  
However, the Pigeon Guillemot in the image actually has a black beak, which is properly localized using our proposed method.

Additionally, we also compare our phrase-critic ranking method to a ranking method based solely on sentences fluency ($S_f$).
We sample 100 random images from the test set and find that attributes mentioned by our critic model reflect the image more accurately than the baseline (85\% image relevant attributes vs. 79\%).

Figure~\ref{fig:CriticVBaseline} shows sampled sentences and their corresponding scores.  We see a precise localization of small regions such as ``white eye'' for ``Brewer Blackbird''. Note that the highest ranked explanations for  ``Brewer Blackbird'' both correctly mention  ``white eye'' which is the strongest distinguishing property of this bird from other black birds. The 3rd sentence gets ranked lower likely due to the explanation ``lighter colored eye-coloring'' lacking fluency. 
Additionally, explanations preferred by our phrase-critic model do not blindly mention class-specific attributes that do not appear in the image.

\small{
\bibliography{nips_2017}
\bibliographystyle{abbrv}
}

\newpage
\appendix

\end{document}

%% file: macros.tex
\newcommand{\Loss}{\mathcal{L}}

%% file: nips_2017.bbl
\begin{thebibliography}{1}

\bibitem{biran2014justification}
O.~Biran and K.~McKeown.
\newblock Justification narratives for individual classifications.
\newblock In {\em Proceedings of the AutoML workshop at ICML}, volume 2014,
  2014.

\bibitem{hendricks16eccv}
L.~A. Hendricks, Z.~Akata, M.~Rohrbach, J.~Donahue, B.~Schiele, and T.~Darrell.
\newblock Generating visual explanations.
\newblock In {\em ECCV}, 2016.

\bibitem{hu2017modeling}
R.~Hu, M.~Rohrbach, J.~Andreas, T.~Darrell, and K.~Saenko.
\newblock Modeling relationships in referential expressions with compositional
  modular networks.
\newblock In {\em CVPR}, 2017.

\bibitem{krishna2016visual}
R.~Krishna, Y.~Zhu, O.~Groth, J.~Johnson, K.~Hata, J.~Kravitz, S.~Chen,
  Y.~Kalantidis, L.-J. Li, D.~A. Shamma, et~al.
\newblock Visual genome: Connecting language and vision using crowdsourced
  dense image annotations.
\newblock {\em arXiv preprint arXiv:1602.07332}, 2016.

\bibitem{relativeattributesiccv}
D.~Parikh and K.~Grauman.
\newblock Relative attributes.
\newblock In {\em ICCV}, 2011.

\bibitem{park2016attentive}
D.~H. Park, L.~A. Hendricks, Z.~Akata, B.~Schiele, T.~Darrell, and M.~Rohrbach.
\newblock Attentive explanations: Justifying decisions and pointing to the
  evidence.
\newblock {\em arXiv preprint arXiv:1612.04757}, 2016.

\bibitem{RALS16}
S.~Reed, Z.~Akata, H.~Lee, and B.~Schiele.
\newblock Learning deep representations of fine-grained visual descriptions.
\newblock In {\em CVPR}, 2016.

\bibitem{teach1981analysis}
R.~L. Teach and E.~H. Shortliffe.
\newblock An analysis of physician attitudes regarding computer-based clinical
  consultation systems.
\newblock {\em Computers and Biomedical Research}, 14(6):542--558, 1981.

\bibitem{welinder10tr}
P.~Welinder, S.~Branson, T.~Mita, C.~Wah, F.~Schroff, S.~Belongie, and
  P.~Perona.
\newblock Caltech-ucsd birds 200.
\newblock Technical report, California Institute of Technology, 2010.

\end{thebibliography}
